\documentclass[10pt,journal,compsoc]{IEEEtran}
\ifCLASSOPTIONcompsoc
  \usepackage[nocompress]{cite}
\else
  \usepackage{cite}
\fi
\ifCLASSINFOpdf
  \usepackage[pdftex]{graphicx}
  \graphicspath{{Figures/}}
\else
  \usepackage[dvips]{graphicx}
\fi
\usepackage{amsmath}
\usepackage{algorithmic}
\usepackage{array}
  \usepackage[caption=false,font=footnotesize,labelfont=sf,textfont=sf]{subfig}
  \usepackage[caption=false,font=footnotesize]{subfig}
\usepackage{fixltx2e}
\usepackage{stfloats}
  \usepackage[nomarkers]{endfloat}
\usepackage{url}
\begin{document}
%
% paper title
% Titles are generally capitalized except for words such as a, an, and, as,
% at, but, by, for, in, nor, of, on, or, the, to and up, which are usually
% not capitalized unless they are the first or last word of the title.
% Linebreaks \\ can be used within to get better formatting as desired.
% Do not put math or special symbols in the title.
\title{A Survey of Deep Learning Techniques for\\ Mobile Robot Applications}
%
%
% author names and IEEE memberships
% note positions of commas and nonbreaking spaces ( ~ ) LaTeX will not break
% a structure at a ~ so this keeps an author's name from being broken across
% two lines.
% use \thanks{} to gain access to the first footnote area
% a separate \thanks must be used for each paragraph as LaTeX2e's \thanks
% was not built to handle multiple paragraphs
%
%
%\IEEEcompsocitemizethanks is a special \thanks that produces the bulleted
% lists the Computer Society journals use for "first footnote" author
% affiliations. Use \IEEEcompsocthanksitem which works much like \item
% for each affiliation group. When not in compsoc mode,
% \IEEEcompsocitemizethanks becomes like \thanks and
% \IEEEcompsocthanksitem becomes a line break with idention. This
% facilitates dual compilation, although admittedly the differences in the
% desired content of \author between the different types of papers makes a
% one-size-fits-all approach a daunting prospect. For instance, compsoc 
% journal papers have the author affiliations above the "Manuscript
% received ..."  text while in non-compsoc journals this is reversed. Sigh.

%\author
%{Jahanzaib~Shabbir,~\IEEEmembership{Member,~IEEE,} and
 %Tarique~Anwer~\IEEEmembership{Fellow,~OSA,}}

\author
{Jahanzaib~Shabbir, and
 Tarique~Anwer}

\markboth{Journal of \LaTeX\ Class Files,~Vol.~14, No.~8, August~2015}%
{Shell \MakeLowercase{\textit{et al.}}: Bare Demo of IEEEtran.cls for Computer Society Journals}
% The only time the second header will appear is for the odd numbered pages
% after the title page when using the twoside option.
% 
% *** Note that you probably will NOT want to include the author's ***
% *** name in the headers of peer review papers.                   ***
% You can use \ifCLASSOPTIONpeerreview for conditional compilation here if
% you desire.

% The publisher's ID mark at the bottom of the page is less important with
% Computer Society journal papers as those publications place the marks
% outside of the main text columns and, therefore, unlike regular IEEE
% journals, the available text space is not reduced by their presence.
% If you want to put a publisher's ID mark on the page you can do it like
% this:
%\IEEEpubid{0000--0000/00\$00.00~\copyright~2015 IEEE}
% or like this to get the Computer Society new two part style.
%\IEEEpubid{\makebox[\columnwidth]{\hfill 0000--0000/00/\$00.00~\copyright~2015 IEEE}%
%\hspace{\columnsep}\makebox[\columnwidth]{Published by the IEEE Computer Society\hfill}}
% Remember, if you use this you must call \IEEEpubidadjcol in the second
% column for its text to clear the IEEEpubid mark (Computer Society jorunal
% papers don't need this extra clearance.)

% use for special paper notices
%\IEEEspecialpapernotice{(Invited Paper)}

% for Computer Society papers, we must declare the abstract and index terms
% PRIOR to the title within the \IEEEtitleabstractindextext IEEEtran
% command as these need to go into the title area created by \maketitle.
% As a general rule, do not put math, special symbols or citations
% in the abstract or keywords.
\IEEEtitleabstractindextext{%
\begin{abstract}
Advancements in deep learning over the years have attracted research into how deep artificial neural networks can be used in robotic systems. It is on this basis that the following research survey will present a discussion of the applications, gains, and obstacles to deep learning in comparison to physical robotic systems while using modern research as examples. The research survey will present a summarization of the current research with specific focus on the gains and obstacles in comparison to robotics. This will be followed by a primer on discussing how notable deep learning structures can be used in robotics with relevant examples. The next section will show the practical considerations robotics researchers desire to use in regard to deep learning neural networks. Finally, the research survey will show the shortcomings and solutions to mitigate them in addition to discussion of the future trends. The intention of this research is to show how recent advancements in the broader robotics field can inspire additional research in applying deep learning in robotics.
\end{abstract}

% Note that keywords are not normally used for peerreview papers.
}

% make the title area
\maketitle

% To allow for easy dual compilation without having to reenter the
% abstract/keywords data, the \IEEEtitleabstractindextext text will
% not be used in maketitle, but will appear (i.e., to be "transported")
% here as \IEEEdisplaynontitleabstractindextext when the compsoc 
% or transmag modes are not selected <OR> if conference mode is selected 
% - because all conference papers position the abstract like regular
% papers do.
\IEEEdisplaynontitleabstractindextext
% \IEEEdisplaynontitleabstractindextext has no effect when using
% compsoc or transmag under a non-conference mode.

% For peer review papers, you can put extra information on the cover
% page as needed:
% \ifCLASSOPTIONpeerreview
% \begin{center} \bfseries EDICS Category: 3-BBND \end{center}
% \fi
%
% For peerreview papers, this IEEEtran command inserts a page break and
% creates the second title. It will be ignored for other modes.
\IEEEpeerreviewmaketitle

\IEEEraisesectionheading{\section{Introduction}\label{sec:introduction}}
% Computer Society journal (but not conference!) papers do something unusual
% with the very first section heading (almost always called "Introduction").
% They place it ABOVE the main text! IEEEtran.cls does not automatically do
% this for you, but you can achieve this effect with the provided
% \IEEEraisesectionheading{} command. Note the need to keep any \label that
% is to refer to the section immediately after \section in the above as
% \IEEEraisesectionheading puts \section within a raised box.

% The very first letter is a 2 line initial drop letter followed
% by the rest of the first word in caps (small caps for compsoc).
% 
% form to use if the first word consists of a single letter:
% \IEEEPARstart{A}{demo} file is ....
% 
% form to use if you need the single drop letter followed by
% normal text (unknown if ever used by the IEEE):
% \IEEEPARstart{A}{}demo file is ....
% 
% Some journals put the first two words in caps:
% \IEEEPARstart{T}{his demo} file is ....
% 
% Here we have the typical use of a "T" for an initial drop letter
% and "HIS" in caps to complete the first word.
%\IEEEPARstart{T}{his} demo file is intended to serve as a ``starter file''
%for IEEE Computer Society journal papers produced under \LaTeX\ using
%IEEEtran.cls version 1.8b and later.
% You must have at least 2 lines in the paragraph with the drop letter
% (should never be an issue)
%I wish you the best of success.

%\hfill mds
 
%\hfill August 26, 2015

\subsection{Defining Deep Learning in the Context of Robotic Systems}
Deep learning is defined as the field of science that involves training extensive artificial neural networks using complex functions, for example, nonlinear dynamics to change data from a raw, high-dimension, multimodal state to that which can be understood by a robotic system \cite{lecun2015deep}. 
However, deep learning entails certain shortcomings which affect physical robotic systems whereby generation of training data in overall is costly and therefore sub-optimal performance in the course of training poses a risk in certain applications. Yet, even with such difficulties, robotics researchers are searching for creative options, for instance, leveraging training data through digital manipulation, automated training and using multiple deep neural networks to improve the performance and lower the time for training \cite{lillicrap2015continuous}.
The idea of using machine learning to control robots needs humans to show the willingness to lose a certain measure of control. This is seemingly counterintuitive in the beginning although the gain for doing this is to allow the system to begin learning on its own \cite{esteves2003generalized}. This makes the systems capable of adaptation such that the potential of ultimately improving their direction is that originating from human control. 
This makes deep neural networks well suited to be used with robots since they are flexible and can be used in frameworks that cannot be supported by other machine learning models \cite{vedaldi2015matconvnet}. For a long time, the most notable method for optimization in neural networks is known as the stochastic gradient descent. However, improved techniques, for instance, RMSProp, as well as Adam of recent, have gained widespread use. Each of the many types of deep learning models is made through the stacking of several layers of regression models \cite{eitel2015multimodal}. Within these models, distinct types of layers have undergone evolution for many aims.
\subsection{Forms of Deep Learning Applied to Mobile Robotic Systems}
One type of layer that demands specific mention is convolutional layers. Unlike traditional layers that are fully connected, convolutional layers apply the same weights in order to operate in all the input space. This brings about a significant reduction of the overall number of weights in the neural network which is specifically vital with images that normally compose of hundreds of thousands and millions of pixels that require processing \cite{yang2015robot}.
 It should be noted that processing these kinds of images which have fully connected layers would need over 100K2 to 1M2 weights which connect to each layer which makes it entirely impractical. The inspiration of convolutional layers came from cortical neurons within the visual cortex which only respond to stimuli in a receptive environment. 
Since convolution estimates such behavior, convolutional layers can be expected to excel at image processing assignments \cite{cirecsan2010deep}. The pioneering research in neural networks using convolutional layers uses image recognition tasks which we built on the advancements of ImageNet recognition competitions around 2012. The lessons learned in this period gained widespread interest in convolutional layers being able to gain super-human recognition of images \cite{lu2015transfer}. 
Currently, convolutional neural networks have been come well known and highly effective as a deep learning model for many image-based applications. These applications comprise of semantic image segmentation, scaling images using super-resolution, scene recognition, object localization with images, human gesture recognition and facial recognition \cite{turan2017deep1} \cite{jordan2015machine}.
Images are not the only form of a signal which illustrates the excellence of convolutional neural networks. Their capability is also effective in any form of a signal which demonstrates spatiotemporal proximity for example speech recognition as well as speech and audio synthesis \cite{turan2018deep}. 
Naturally, these have also started to be dominative in the domain of signal processing and heavily used in robotics, for instance, pedestrian detection with the use of LIDAR, mico-Doppler signatures, as well as depth-map, estimating \cite{sunderhauf2015place} \cite{turan2017fully}. Recent projects have even started to integrate signals from several modalities and combine them for unified recognition and perception \cite{turan2018sparse}.
Ultimately, the philosophy that underlies and prevails in the deep learning community is that every component of a complex system can be taught to "learn." Therefore, the actual power of deep learning does not come from applying just one of the described structures in the previous section as a part in robotics systems by in connecting components of all these structures to form a complete system that learns entirely \cite{lake2015human}.
 This is the point where deep learning starts to make its impact such that each component of the system is capable of learning as a whole and is capable of adapting to sophisticated methods. For example, neuroscientists have even started recognizing the many patterns evolving in the deep learning community and in all artificial intelligence are starting to mirror patterns previously evolved in the human brain \cite{marsland2009machine}.
In the process of learning complex, high-dimensional as well as novel dynamics, the analysis of derivatives within these complex dynamics needs human expertise. However, this process normally consumes a lot of time and can bring about a trade-off between the dimensionality and tractability of states \cite{lenz2015deep}. 
Therefore making these models robust to unforeseen impact is challenging and in most cases, full state information is normally unknown.
 In this case, systems that are able to rapidly and autonomously adapt to modern dynamics are required to solve problems for instance moving over services with unknown or uncertain attributes, managing interaction in a new environment or adapting or degrading robot subsystems \cite{ghahramani2015probabilistic}.
 Therefore, we need methods that are able to accomplish possession of hundreds or thousands of degrees of freedom and demonstrate high measures of uncertainty which are only available in a state of partial information. 
On the other hand, the process of learning control policies in dynamic environments and dynamic control systems is able to accommodate high measures of freedom for applications such as swarm robotics, anthropomorphic hands, robot vision, autonomous robot driving and robotic arm manipulation \cite{duan2016benchmarking}. 
However, despite the advancements gained over the years in active research, robust and overall solutions for tasks, for instance, moving in deformed surfaces or navigating complex geometries with the use of tools and actuator systems have remained elusive more so in novel scenarios. This shortcoming is inclusive of kinematic and path planning tasks which are inherent in advanced movement \cite{levine2016learning}.
On the other hand, in terms of advanced object recognition, deep neural networks have proved to be increasingly adapt to the recognition and classification of objects. Examples of advanced application include recognition of deformed objects and estimation of their state and pose for movement, semantic task, and path specification, for example, moving around the table \cite{levine2016learning}. In addition, it includes recognizing the attributes of an object and surface whereby for instance a sharp object could present a danger to human collaborators in certain environments such as rough terrain.
In the face of such difficulties, deep learning models can be used in the approximation of functions from sample input-output pairs. These can be the most general purpose deep learning structures since there are several distinct functions in robotics which researchers can use in approximating from sample observations \cite{yang2015grasp}.
 Certain examples of these observations entail mapping from actions to corresponding changes in the stage, mapping these changes in state to actions that can cause it or mapping from force to motion. 
While in certain cases particular physical equations for such functions may already be defined, there are several situations where the environment is highly complex for such equations to generate acceptable accuracy \cite{dong2016image}. 
However, in such scenarios learning approximation of functions from sample observations can yield accuracy that is significantly better.  In other words, approximated functions do not need to be continuous. However, function approximation models are also excellent at classification tasks, for instance, determining the type of obstacles before a robot, the overall path planning strategy well suited for present environments or the state of a certain complex object which the object is interacting with \cite{gongal2015sensors}.
Furthermore, function approximation deep learning architecture using rectifiers can model the high coupled dynamics of an autonomous mobile robot to solve the analytic derivatives and challenging system identification problems.
 Deep neural networks have superseded other models in detection and perception since they are capable of engaging in direct operation with highly-dimensional input rather than needing feature vectors based on hand-engineered designs by humans \cite{nguyen2015innovation}. This lowers the dependence on humans such that additional training time can be partially offset by lowering initial engineering efforts \cite{du2015hierarchical}. 
Extraction of meaning from video or still imagery is another application where deep learning has gained impressive progression. This process demands simultaneously addressing using the four independent factors of object detection and a single deep neural network. These factors include feature extraction, motion handling, classification, articulation and occlusion handling \cite{tang2013deep}. 
Unified systems limit suboptimal interactions between normally separate systems by predicting the physical results of dynamic scenes using vision alone. This is based on the actions of humans to be able to predict the results of a dynamic scene from visual information, for example, a rock falling down and impacting another rock \cite{sunderhauf2015performance}.
 It is therefore on this premise that deep learning has been identified as being effective in managing multimodal data generation in robotic sensor applications. These applications include integration of vision and haptic sensor data, incorporating depth data and image information from RGB-D camera data. Due to the extensive number of meta-parameters, deep neural networks have evolved somewhat a reputation of being challenging for non-experts to be used effectively \cite{chen2015deepdriving}. 
However, such parameters also avail significant flexibility which is a vital factor in their general success. Therefore, training deep neural networks needs the user to be able to develop at least an elementary level of familiarization with many concepts. Specifically, applying these techniques will help in tacking advanced object recognition challenges and reduced the extent of the entire changes as well \cite{schmidhuber2015deep}.

\section{Deep Learning for Robotic Perception}

\subsection{Current Robotic Perception Trends}

Although current trends are more leaning to deep and big models, a simplified neural network with just a single hidden layer and a basic sigmoid shaped activation function will train faster and provide a baseline that is used to give meaning to any deeper model improvements. 
When we use deeper models, Leaky Rectifiers are able to normally promote faster training by lowering the impact of the diminishing gradient challenge and improving accuracy through using simplified monotonic derivatives \cite{brighton2015introducing}. 
Furthermore, since models with additional weights have increased flexibility to over fitting training data, regularization is a vital technique in training the best model. In addition, an elastic net is a combination of well-established regularization methods used in promoting robustness against weight saturation and also promotion sparsity in weights \cite{bai2015subset}. However, newer regularization methods inclusive of drop-out and drop-connect has attained even better empirical outcomes. 
Furthermore, many regularization methods are also in existence in specifically improving the robustness of autoencoders. In this case, special-purpose layers can also make a significant distinction with deep neural networks \cite{veeriah2015differential}. It is a common method to alternate between convolutional and max pool layers. These pool layers can lower the general number of weights in the network and also allow the model to be able to recognize objects independent of where they are placed in the visual field \cite{xu2015jointly}.
On the other hand, batch normalization can provide us with significant improvements in rating the convergence by ensuring the gradient in range affects the weights of all neurons. In addition, residual layers can allow a deeper and consequently more flexible model to be trained \cite{narasimhan2015language}. 
To make effective use of deep learning models, it is vital to train one or many General Purpose Graphical Processing Units since the other methods of parallelization of deep neural networks have been tried but none of them have yet provided gains in beneficial performance of General Purpose Graphical Processing Units \cite{wu2015galileo}.
For a long time until in recent years, robots have long been used in industrial environments. 
In industrial environments, robotic systems are pre-programmed with repetitive assignments which lack the capability of autonomy and as such operate on the basis of a structured approach. Such an environment cannot be adaptive for a mobile robot since it eliminates the need for autonomy. On the other hand, mobile robots need less structured environments such that they can be able to make their own decisions such as navigate paths, determine whether objects are obstacles, recognize images and audio as well as map their environments. As such, surviving and adapting in the real world is more complex for any robotic system in comparison to the industrial setting since the risk of failure, system error, external factors, obstacles, corrupt data, human error and unrecognizable environments is more prevalent \cite{lake2015human}.

\subsection{Machine Learning Usage in Training Robotic System Perception}

The difference between deep learning and machine learning is that deep learning place emphasis on the subset of machine learning resources and method and uses them to solve any difficulties that need "thought" whether human or artificial. Deep learning is also introduced as a means of making sense of data with the use of multiple abstraction layers. In the course of the training process, deep neural networks are able to learn the means of discovering useful patterns to digitally represent data such as sounds and images. This is specifically why we observe more advances in the areas of image recognition and natural language processing originating from deep learning \cite{turan2018unsupervised}.
It is on this backdrop that deep learning has taken the forefront position in helping researchers develop breakthrough methods to the perception capabilities of robotics systems \cite{marsland2009machine}.
In more simplified language, perception refers to the functionality of robots being able to detect its surroundings. It is therefore heavily reliant on multiple sources of sensory information. However, with traditional robot technology extracting data from raw sensor by using rudimentary constructed sensors theseold methods were limited by constraints of adapting to generic settings \cite{lenz2015deep}.
In situations where these robotic systems faced dynamic environments, they operated in an unstructured manner by combining hybrid and autonomous functionality to process information about their surroundings \cite{ghahramani2015probabilistic}.  
As such, with deep learning came the introduction of new methods of processing data from robotic sensorsof a robotic system’s surroundings using a feature known as perception. These methods comprise of robot motion, perception, human-interaction, manipulation and grasping, automation, self-supervision, self training and learning as well as robot vision \cite{duan2016benchmarking}. 
Deep learning models utilize automated actions technology at only half the cost by using supervised learning to attain their goals \cite{levine2016learning}. For example, in order to perfect an image recognition application, a neural net will be required to be trained with a collection of labeled data. 
On the other hand, unsupervised learning is how deep learning operates and allows for the discovery of new patterns and insights by tackling problems with little or no insight on what the results should be perceived by a robot \cite{yang2015grasp}. The method by which a mobile robot is able to detect it’s environment with the use perception is by using definitive decision-making policies \cite{levine2016learning}. For instance, mobile robots using deep learning are able to navigate with rationality by using motion and track precision sensors which are driven machine learning algorithms \cite{yang2015grasp}. However, in difficult environments such as a congested room, the level of accurately perceiving their environment is limited. Therefore, deep learning based solutions can tackle this challenge by using artificial intelligence, high computational hardware and processing layers known as deep convolution neural networks to solve this dilemma by successfully deciphering intricate environment perception difficulties \cite{dong2016image}.
The various obstacles that exist in a robot’s environment are an indication of the immense high-dimensional data processing capabilitiesrequiredby a robotic system to be able to perceive its surroundings \cite{gongal2015sensors}. 
By using self-supervised, semi-supervised and full supervised training coupled with learning, robotic systems are able to utilize their machine learning and pattern recognition capabilities to process raw data such as images, objects, and semantics, audio as well as process natural language in the real time. 
This segment of deep learning is the best since it uses feed-forward artificial neural networks to successfully analyze visual imagery. It also uses multiple multilayer perceptrons based on designs that need low preprocessing \cite{dong2016image}. 
In comparison to other image classification algorithms, it uses minimal pre-processing which means that the network is able to learn filters using automated procedures unlike traditional algorithms that are manually engineered. Therefore, by not relying on past knowledge and human efforts in the design of features makes it a major advantage \cite{nguyen2015innovation}.
As such in general, this makes deep learning capable of the extraction of multi-level attributes from raw sensor data in a direct manner with no need for human assisted robotic control \cite{nguyen2015innovation}. 
This presents researchers with the implication that deep learning programming librariessuch as TensorflowTheano of Python, Caffe of C++, darch in R, CNTK, Convent.js of Javascript, and Deeplearning4j derived from C++ and Java among others are extremely of use in providing robotic systems with a platform to develop their sensory data analysis and environmentlearningby using deep learning algorithms \cite{du2015hierarchical}.
Robotic system perception concerns auxiliary functions within mobile robotic are vital in interactingwith a robot’s environment.Sensing and intelligent perception are some of the applications which are vital since they determine the performance of a robotic system. 
These performances are largely dependent on how robot sensors perform. Modern sensors and their functionality can provide impressive robot perception which is the foundation of self-adaptive as well as robotic artificial intelligence \cite{tang2013deep}. 
The process of changing from sensory input to control output using sensory-motor control structures presentsa big difficultyin robotic perception \cite{sunderhauf2015performance}. Some of the vital mobile robot components include the manipulator comprising of many joints and connections, a locomotion device, sensors, a controller and an endeffector \cite{turan2017endosensorfusion}. Mobile robots are automated systems capable of moving. They also have the ability to move around their surroundings and are not fixed to a single physical location.
With these features mobile robots are able to perceive their environments usingsensory data andautonomous control commands \cite{chen2015deepdriving}. 

\subsection{Shortcomings of Deep Learning in Robotic Perception}

However, certain challenges remain unresolved in these robotic systems particularly the areas of perception and intelligent control.
Some of these challenges are reflected in the process needing a lot of data to be able to train and teach algorithms progressively. Large datasets are required to ensure machines deliver the desired outcomes. In the same way as the human brain needs rich experience to learn and deduce information, artificial neural networks also need abundant amounts of data. This means for more powerful abstractions, more parameters are required and hence more data. 
Another challenge is the tendency of over fitting in neural networks whereby in certain cases, there is a sharp distinction between an error within a training set and that encountered into new untrained datasets. This problem arises when many models make the relative number of parameters fail to reliably perform. Therefore, the model only memorizes training examples and fails to learn generalization of new situations and new datasets.

\subsection{How Mobile Robotic Perception Models Can Be Used to Attain Complete Situational Awareness}

These capabilities will be surveyed within the research survey through deep learning perception model algorithms that are ableto determine how a robot responds to the dynamic changes within an environment \cite{schmidhuber2015deep}. The basis of these models revolves around control theory affiliated paradigms for instance system stability, control as well as observation \cite{brighton2015introducing}.
This theory states that a robotic system is able to perceive its environment by using hierarchical extensions or enhancements of learning by maximizing the range of its sensor capabilities while using path planning algorithms to maneuver around obstacles or paths \cite{bai2015subset}.
Most real-time map algorithms are concerned with the acquisition of compact 3D mapping within indoor settings with the use of range as well as imaging sensory capabilities \cite{veeriah2015differential}.  The process of developing models of a robot's environment is a vital problem to deal with more so in regard to managing its workspace especially when it is shared with other machinery \cite{xu2015jointly}. A mobile robot interacts with the environment by using control systems which define structures or obstacles as geometrical areas so as to be able to cover all the likely configurations on the robot. Objects or structures are defined according to parallelepipeds, spheres, planes, and cylinders. With such a simplified model, the mobile robot system is able to define many geometrical areas of this nature to cover nearly all objects within its surrounding, for example, moving objects, stationary items such as furniture and machinery. Therefore, we propose elementary geometrical volumes as a means of modeling a mobile robot's perception capabilities of its environment  \cite{narasimhan2015language}.
 This method will allow the robot to be able to move within an environment with the certainty of not colliding with regions that are forbidden; these regions must already be defined, declared and activated so as to be able to correctly work \cite{wu2015galileo}.
% An example of a floating figure using the graphicx package.
% Note that \label must occur AFTER (or within) \caption.
% For figures, \caption should occur after the \includegraphics.
% Note that IEEEtran v1.7 and later has special internal code that
% is designed to preserve the operation of \label within \caption
% even when the captionsoff option is in effect. However, because
% of issues like this, it may be the safest practice to put all your
% \label just after \caption rather than within \caption{}.
%
% Reminder: the "draftcls" or "draftclsnofoot", not "draft", class
% option should be used if it is desired that the figures are to be
% displayed while in draft mode.
%

The geometrical perception algorithm will check if the robot end effector is within the controlled area or warning zone. This checking is done through the use of already stored geometrical areas already defined by the user \cite{lun2015survey}.
 Similarly, in this context, the position of the dynamic object avails the perception system with the likelihood of connecting the geometrical regions with arbitrary moving points \cite{kuen2015self}.
 These points can be read from external sensors such as encoders. Control of speed is undertaken with the user of geometrical area blocks able to detect the shape typology and choose the correct movement law to be used so as to modify the robot override and avoid collisions with user-defined zones \cite{hou2015convolutional}.
The speed override is transformed smoothly when the robot end effector collides with a spherical zone in accordance with the perception law in \eqref{eq1} \cite{esteves2003generalized}.
\begin{equation}\label{eq1}
\begin{split}
V=v_0 . \frac{d-r}{\delta}, r \leq d \leq r+\delta, \\ V=v_0,  d > r+\delta, and \\ V=0, d<r
\end{split}
\end{equation} 

In this case, $v$ is represented as the robot end effector actual speed override, $v_0$ as the old override, $d$ as the distance between the robot end-effector and the main spherical area, the thickness of the warning zone is represented as $\delta$ while $r$ is the sphere radius \cite{qian2015deep}.
 At thestage when the robot meets a cylindrical zone, the speed override will be subjected to the perception law in \eqref{eq2} \cite{esteves2003generalized}.

\begin{equation}\label{eq2}
\begin{split}
V=v_0 . \frac{d_1-r}{R-r}, 0 \leq z \leq h, \\ V=v_0 . \frac{d_2}{\delta}, h \leq z \leq h+\delta, -\delta \leq z < 0, p  \in  cyl, and \\ V=v_0 . \frac{d_3}{\delta}, h \leq z \leq h+\delta, -\delta \leq z < 0, p \in cyl
\end{split}
\end{equation} 

The perception law above represents $h$ as the cylinder height, the position of the robot end effector is denoted as $p$, the distance between the center of the cylinder and the position of the robot is denoted as $d_1$ while the $d_2$ represents the distance between the top or bottom base of the cylinder as well as the position of the robot \cite{tzeng2015simultaneous}. 

In addition, the least position between the position of the robot and the top/bottom circumference points of the cylinder base is denoted as $d_3$. Furthermore, the robot speed override coincides with the past speed override at the stage when the robot end effector is outside the warning zone \cite{pinto2016curious}.

\section{Deep Learning for Robotic Control and Exploration}

\subsection{How Autonomous Robotic Systems Use Deep Learning to Control and Explore Their Enivornments}

Realizing the benefits of autonomous robot exploration presents robotics researchers with many applications of considerable community and financial impact \cite{lake2017building}. 
Robotics research relies on perfect knowledge and control of the environment \cite{chen2015combining}. The problems related to unstructured environments are an outcome of the high-dimensional state space as well as the inherent likelihood in mapping sensory perceptions on particular states. It should be noted that the high dimensionality of the state space is representative of the most basic difficulty since robots leave highly controlled environments of a laboratory and enter into unstructured surroundings. For example autonomous unmanned aerial vehicles used deep learning to classify terrain and solve any exploration shortcomings by generating control commands for its human operator so as to adapt to a certain trade-off \cite{wulff2015efficient}. 
The major hypothesis of this approach is therefore for mobile robots to succeed in unstructured surroundings such that they can carefully choose assignment specific attributes and identify the relevant real-time structures to lower their state space without impacting the performance of their exploration objectives \cite{ruiz2015scene}.
Robots perform assignments by exploring their surroundings. As such, given our focus on autonomous mobile exploration, we shall direct most attention to exploration in service of movement, that is to say collision-free movement for end-effector placement \cite{das2015performance}. The challenge of generating such movement is an example of the problem faced in motion planning. Motion planning for robotic systems with many levels of freedom is computationally challenging even in environments that are highly structured due to the increased-dimensional configuration space \cite{salakhutdinov2015learning}.
In addition, unstructured environments are associated with the imposition of added difficulties in motion generation in comparison to the traditional motion planning process. Furthermore, in environments that are unstructured, a robot is only able to possess limited knowledge of its environment, objects are able to change their unknown to a known state for the robot by manipulating assignments may using the end effector \cite{schmidhuber2015learning}. However, this capability can be challenged by to a constrained trajectory such that the mobile robot is unable to reach a particular location with ease \cite{turan2017non1}.
Each of these problems makes the challenge of motion generation more complex. The explicit coordination of planning and sensing necessary to manage dynamic environments increases the dimension of the state space. In addition, robotic assignment requirements impose stringent terms in form of high-frequency feedback \cite{chen2015road}.
Therefore, existing motion planners tend to make assumptions that are highly restrictive for environments that are unstructured since they are highly computationally difficult to satisfy in terms of gaining operator-to-robot feedback \cite{vrigkas2015review}. These assumptions, as well as the computational difficulty, are an outcome of the foundation of motion planning with its high-dimensional configuration spacing which makes it highly unsuited in solving space problems.
 Planners instead can be able to solve this paradigm by solely using workspace information for collision avoidance. Nearly all real-world surroundings, however, comprised of a considerable measure of structure \cite{oyedotun2017deep}. For instance, buildings are segmented into hallways, doors, and rooms; outdoor environments comprise of paths, streets, and intersections while objects for example tables, shelves and chairs have more favorable approach directions. This information, however, is neglected when robot exploration planners exclusively operate on the basis of a configuration space \cite{moore2015talking}. Therefore, the outcome for most robotic exploration planners is to operate by the assumption that any environment is perfectly defined and sustained as static in the course of planning \cite{hinton2006fast}.

\section{Deep Learning in Robotic Robotic Navigation and Autonomous Driving}

Using deep learning to attain autonomous driving assignment is not a perfectly controlled and modeled task as most people think. Instead, it needs optimal perceptual capabilities \cite{zhu2017target}. The process of perception of a robot’s environment and interpreting the information it acquires allows it to understand the condition of its surroundings, devise plans to change the state and observe how its actions impact its environment. In unstructured environments, recognition of objects has been proven to be highly challenging. With immense volumes of sensor data and increased variation of objects within similar object categories, for example, a paved and unpaved road as well as recognizing objects \cite{cruz2015interactive}. 
Deep learning uses machine learning motion capturing abilities as well as optimal perceptual functionalities. This is a prerequisite of several vital applications for robots for instance flexible manufacturing, planetary exploration, collaboration with human experts and elder care \cite{vinciarelli2015open}.
The challenge of driving in an environment includes problems of movement of the robot in navigating varied obstacles by pushing and pulling. Even in structured environments, automated driving is difficult due to the complexities of the related state space \cite{doshi2015deep}. This state space comprises of appearance, dimension, position as well as the weight of objects within the scene. It also comprises of several other relevant attributes which provide indications of where to pull, push or grasp as well as the level of force to apply \cite{wang2015designing}. Deep learning and associated machine learning collective actions improved the performance of robots in making the decision by embedding the capability within these systems to choose between possible actions and determine the required parameters for their controllers.

\subsection{How Autonomous Robotic Systems Use Deep Learning to Navigate Their Enivornments}

Autonomous driving in unstructured environments faces many challenges which do not exist in structured environments. In unstructured environments, object attributes needed for driving cannot be defined as priori. Information concerning objects has to be gained through sensors even though these are normally ambiguous and therefore introduce uncertainty and avail information that is redundant. Furthermore, autonomous driving in unstructured and dynamic surroundings normally needs responding in a fashion that is timely to a rapidly transforming environment \cite{cuayahuitl2015strategic}.
Problems of autonomous driving can be made simple through the exploitation of the structure which is inherent to human surroundings. Most objects in the real world are based on designs to perform certain functions with the intention of being utilized by humans \cite{lake2017building}. As an outcome, several real-world objects can share common attributes which allude to their intended usage. By placing emphasis on these assignment-related object attributes, the complexity of autonomous driving is lowered.
For example, visual data can be analyzed for the identification of few points which correspond to positive locations at which a robot can maneuver. Furthermore, since movement features are similar across many objects, robots could be trained to identify them. As a consequence, the state space that requires exploration in order to move is significantly lowered \cite{ohn2016looking}.
In this case, researchers normally make assumptions to lower the complexity of autonomous driving in unstructured environments. For instance, it is normally assumed that full models of objects in the environment can be availed a priori or it can be gained through sensors while the environment remains the same in the process of interaction \cite{wei2017robotic}. 
However, in practical terms, it is impossible to avail autonomous driving with full priori models in the actual world. However, models that are perfect are not a prerequisite for successful autonomous driving \cite{mathieu2015deep}. Robot mobile driving can be guided using existing structures in the world and in most cases those which are easy to perceive. 
As such, by leveraging this structure, the complexity of autonomous driving in unstructured environments is lowered significantly. Similarly, understanding the intrinsic measures of freedom of objects in an environment is also able to lower the complexity of autonomous driving in unstructured surroundings \cite{chen2015combining}.

\section{Semi-Supervised and Self-Supervised Learning for Robotics}

Imitation based learning is a promising approach to tackling the difficult robotic assignments, for instance, autonomous navigation. However, it needs human supervision to oversee the process of training and sending correct control commands to robots without feedback. It is this form of procedure that is prone to failure and high cost \cite{szegedy2016rethinking}. Therefore, in order to lower human involvement and limit manual data labeling of autonomous robotic navigation using imitation learning, the techniques of semi-supervised and self-supervised learning can be introduced. 
It should be noted however that these techniques need to operate according to a multi-sensory design approach. The solution should comprise of a suboptimal sensor policy founded on sensor fusion and automatic labeling of states the robot could encounter \cite{turan2017deep}. This is also aimed at eliminating human supervision in the course of the learning process \cite{tang2016extreme} \cite{turan2017non}.
 Furthermore, a recording policy needs to be developed to provide throttling of the adversarial impact of too much data being learned from the suboptimal sensor policy. As such, this solution will equip the robot with the capability of achieving near-human performance to a large extent in most of its assignments \cite{chen2016deeplab}. It is also capable of surpassing human performance in situations of unexpected outcomes for instance hardware failure or human operator error. 
Furthermore, the semi-supervised method can be considered as a solution to the problem of track classification in congested environments such as a room. This problem entails object classification undergoing segmenting and tracking without using class models \cite{chen2016unconstrained}.  Therefore, we introduce semi-supervised learning as a technique capable of solving this problem by iteratively training a classifier and extracting vital training examples from the data in its unlabeled state. This is achieved by exploiting the tracking data. In addition, the process also involves evaluating large multiclass difficulties presented by data sourced from congested artificial natural environments such as a street \cite{papernot2016limitations}.
As such, when provided with manually labeled training tracks of individual object classes, then semi-supervised learning performance in comparison to self-supervised learning is able to use thousands of training tracks \cite{levine2016end}. In addition, when also provided with augmented unlabeled data, semi-supervised learning has demonstrated the capability of outperforming the self-supervised learning method. In this case, semi-supervised learning presents itself as the most simplified algorithmic approach to speeding up incremental updating of booster classifiers by lowering the learning time factor by three \cite{huang2016deep}.

\section{Multimodal Deep Learning Methods for Robotic Vision and Control}

Performing tasks in an imperfect controlled and modeled surrounding means robots need to have optimal multimodal capabilities. The procedure of perceiving the environment and interpreting the gained information allows robots to perceive the state of the environment, devise methods to alter the state and observe the impacts of their actions on the environment %%%%%%%%%[90]. 

\subsection{How Autonomous Robotic Systems Use Semi-Supervised and Self-Supervised Learning to Learn Their Enivornments}

The environment of a robot can be controlled too many levels. In principle, less constrained environments are more difficult to perceive. In the real-world and its unstructured and dynamic surroundings such as vegetation landscape and terrain, the perception of a mobile robot needs to be capable of navigating this unknown environment by using sensor modalities. More so, even without the introduction of uncertainty, sensors in themselves are ambiguous \cite{niekum2015learning}. For example, a lemon and a soccer ball can look similar from a certain perspective. In addition, a cup could be invisible in case the cupboard is shut and it can be challenging to tell the difference between a remote control and cell phone is they are both facing down. These factors are all contributive to the challenges of perceiving the state of the environment. Furthermore, for example, advances in face recognition normally operate under the assumption concerning the position and orientation of the individual in the image. The outcomes of object segmentation are normally founded on the capability of telling the difference between an object and background on the basis of differences in color \cite{devin2017learning}. 
In addition, object recognition is normally reduced to similarities in computing to a limited collection of given objects. On the other hand, in an unstructured environment, the position and orientation are uncontrollable since assumptions concerning color and shades and problematic to justify. Furthermore, the range and likely objects the robot could encounter are intractable \cite{finn2016deep}.

\subsection{How Autonomous Robotic Systems Use Multimodal and Deep Learning Methods to Percieve Their Enivornments}

Therefore, in order to tackle perception in unstructured surroundings, robots need to be able to lower the state space that requires being analyzed. to provide facilitation of certain perceptual assignments by limiting uncertainty and as such lowering the dimensionality of the state space. For instance, in order to compute the distance of objects in an environment, robots need to relate depth to visual information \cite{rusu2016sim}.
 This is normally done with the use of a stereo vision system and solving the correspondence challenge between two static 2D images. In addressing the correspondence problem, however, it is complicated due to noise, many likely matches and the uncertainty in calibrating the camera \cite{mohamed2015variational}.
On the other hand, in a system capable of the capture of at least three view angles in one image, this lowers the state space by reducing a multi-sensor system to a single sensor. Furthermore, in an unstructured environment, recognition of objects has proven to be highly challenging \cite{zhu2017target}.
 This is due to large volumes of sensor data and an increased variation within objects of a similar category. In this case, object recognition is an increased dimensional challenge. However, even in the face of these challenges, objects in the similar category do share similar attributes. 
As such, by applying this insight, robots are able to place emphasis to only a minimal subset of the state space that comprises of the most relevant characteristics for classification \cite{cruz2015interactive}. 

\section{Application of Deep Models to Problems in Vision and Robotics}

The preceding overview of machine learning applications in robotics will highlight five major areas where considerable impacts have been made by robotic technologies currently and in the development levels for long-term use. However, by no means inclusive, the aim of this summarization is to provide the reader with a preview of the form of machine learning applications in existence within robotics and motivate the desire for extended research in such and other fields \cite{maturana2015voxnet}. 
The growth of big data which is to day visual information provided on the internet with the inclusion of annotated images and video has pushed forward advancements in computer vision which has in turn assisted in extending machine-based learning systems to prediction learning methods such as those presented by research at Carnegie Mellon \cite{zhang2015improving}. This presentation involved unveiling offshoot examples such as the anomaly detection using supervised learning have been applied in building structures with the capability of searching and assessing damages in silicon wafers with the use of convolutional neural networks \cite{zhang2015improving}. In addition, extrasensory technologies for instance lidar, ultrasound, and radar such as those developed by Nvidia as also propelling the creation of 360-degree vision-based systems for autonomous vehicles and UAVs \cite{turan2018magnetic}.
Imitation learning which is closely associated with observational learning is also a field categorized by reinforcement learning or the difficulties of gaining an agent to act towards maximizing rewards. One example is Bayesian or probabilistic models which stand out as a common machine learning method used an integral component of field robotics where attributes of mobility in fields such as construction, rescue, and inverse optimal control methods have been utilized in humanoid robotics, off-road terrain navigation, and legged locomotion \cite{finn2017deep}.
Self-supervised learning is another method that allows robots to generate personalized training instances so as to refine performance. This has been integrated robots and optical devices for instance in detection and rejection of objects such as dust and snow, identification of obstacles, vehicle dynamics modeling, and 3D-scene analysis \cite{campos2015diving}.
Assistive and medical technologies are another application where assistive robots entail devices capable of sensing, processing sensory data and undertaking actions that gain individuals with disabilities. Even as smart assistive technologies are existent for the overall population, for instance, driver assistance resources, movement therapy robots avail diagnostic or therapeutic gains.
Furthermore, multi-agent learning concerning coordination and negotiation are vital components involving machine learning based robots also known as agents. This method is broadly utilized in games that are capable of adapting to a transforming landscape of other robots or agents and searching for equilibrium strategies \cite{devin2017learning}.

The interdisciplinary arena of computer vision concerned with how computers are able to be developed for attaining an increased measure of perception from the digital imagery of video. Computer vision assignments comprise of techniques for acquisition, processing, analysis, perceiving digital imagery as well as extracting high dimensional information from the actual world so as to yield numerical or symbolic data for example in the format of decisions \cite{gu2017deep}. 
Artificial intelligence areas concerned with autonomous planning or deliberated robotic systems navigation demand a thorough perception of such settings since information concerning the environment can be availed by a computer vision system in action as a vision sensor \cite{zhu2017target}. Therefore, artificial intelligence, as well as computer vision, share other fields, for instance, pattern recognition and learning methods. 
The consequence is that in certain cases, computer vision is viewed as a component of artificial intelligence. Another application of computer vision is solid state physics since a large portion of computer vision systems are reliant on imagery sensors that provide detection of electromagnetic radiation which normally takes a form that is either visible or infra-red lighting \cite{zhang2015improving} \cite{turan2017endo}. These sensors operate according to quantum physics designs with the procedure by which light interacts with the surface better explained by optics behaviors. 
Such intricate inner working demonstrates how even complex image sensors even need quantum mechanics to avail a full understanding of the process of image formation. Furthermore, another application of computer vision is the multiple measurement challenges in physics which can be tackled by utilizing computer vision for instance fluids motion \cite{kuen2015self}.

\section{Benefits and Drawbacks of Deep Learning to be Applied in Mobile Robots}

\subsection{Benefits of Deep Learning in the Context of Mobile Robots}

The gains of deep learning as a component of the wider family of machine learning techniques founded on representations of learning data in opposition to assignment-particular algorithms through supervised, unsupervised and semi-supervised learning allows for structured on the interpretation of information processing and patterns of communications which can be viewed as trials at defining a relation between multiple stimuli and related neuronal responses \cite{sanchez2016comparative}. 
Deep learning architectures for instance deep neural, deep beliefs as well as recurrent neural networks have been utilized in arenas inclusive of computer vision, natural language processing, social network filtering, speech recognition, bioinformatics and audio recognition.
 In these mentioned fields, deep learning architecture has produced outcomes in comparison to an in certain case more advanced to human expertise. Furthermore, deep learning algorithms utilize a cascade of many nonlinear processing unit layers for extraction of features and transformation with particular layers applying the output from the past layer as input \cite{abadi2016tensorflow}. 

Deep reinforcement learning proposes a simplified conceptual light framework that utilizes asynchronous gradient descent to cater for deep neural network controller optimization. The presented asynchronous variants in standard reinforcement learning algorithms reveal that parallel actor learner holds a stabilizing influence on training which allows for the successful training of the neural network controller \cite{duan2016benchmarking}. It is on this premise that asynchronous variants are presented as the most appealing Deep reinforcement learning approach. 

\subsection{Drawbacks of Deep Learning in the Context of Mobile Robots}

The drawbacks of deep learning in applied robotics is that the storage of the agent data using replay memory does not allow for re-batching or sampling at randomly from varied time-stages. As such, memory aggregation in this approach lowers non-stationary and eroded updates while in simultaneously limiting the techniques to off-policy reinforcement learning algorithms \cite{gongal2015sensors}.

\section{Conclusion}

Deep learning is set to transform the arena of artificial intelligence as well as represent a measure in the direction of developing autonomous systems with an increased scope of perceiving the visual world. Presently, deep learning is allowing scaling of challenges that were traditionally intractable for instance learning to directly play video games for pixels. Furthermore, deep learning algorithms are also utilized in robotics to foster the capability of control functionality for robots indirectly learning from cameral inputs in the actual world. It is on this premise that the survey above illustrates the major advances and approaches of reinforcement learning in regard to the main streams of value and policy-driven methods as well as associated coverage of central algorithms in deep learning for instance deep network, asynchronous advantage actor-critic as well as trust region policy optimization.
 Furthermore, the research survey has highlighted the gains of deep neural networking with emphasis on visual perception through deep learning. One of the core objectives of the discipline of artificial intelligence it the production of completely autonomous agents that are able to interact with their settings to learn optimal behavior and demonstrate improvements with time by trial and error regiments. It is therefore on this premise that creating artificial intelligence systems that are responsive and with the capability of learning has long been an elusive challenge. However, hope is found in the principled mathematical framework of deep learning with utilizes experience driven autonomous learning to apply a functional approximation to represent learning attributes of deep neural networks to overcome these challenges.

\bibliographystyle{IEEEtran}
\bibliography{mybibfile}

% biography section
% 
% If you have an EPS/PDF photo (graphicx package needed) extra braces are
% needed around the contents of the optional argument to biography to prevent
% the LaTeX parser from getting confused when it sees the complicated
% \includegraphics command within an optional argument. (You could create
% your own custom macro containing the \includegraphics command to make things
% simpler here.)
%\begin{IEEEbiography}[{\includegraphics[width=1in,height=1.25in,clip,keepaspectratio]{mshell}}]{Michael Shell}
% or if you just want to reserve a space for a photo:

%\begin{IEEEbiography}{Michael Shell}
%Biography text here.
%\end{IEEEbiography}

%% if you will not have a photo at all:
%\begin{IEEEbiographynophoto}{John Doe}
%Biography text here.
%\end{IEEEbiographynophoto}
%
%% insert where needed to balance the two columns on the last page with
%% biographies
%%\newpage
%
%\begin{IEEEbiographynophoto}{Jane Doe}
%Biography text here.
%\end{IEEEbiographynophoto}
%
%% You can push biographies down or up by placing
%% a \vfill before or after them. The appropriate
%% use of \vfill depends on what kind of text is
%% on the last page and whether or not the columns
%% are being equalized.
%
%%\vfill
%
%% Can be used to pull up biographies so that the bottom of the last one
%% is flush with the other column.
%\enlargethispage{-5in}

% that's all folks
\end{document}